\def\year{2020}\relax
\documentclass[letterpaper]{article} 
\usepackage{aaai20}  
\usepackage{times}  
\usepackage{helvet}  
\usepackage{courier}  
\usepackage{url}  
\usepackage{graphicx}  
\usepackage{makecell}
\usepackage{footnote}
\usepackage{tablefootnote}
\usepackage{enumerate}
\usepackage{amsmath}
\usepackage{amssymb}
\usepackage{algorithmicx}
\usepackage{algorithm}
\usepackage{algpseudocode}
\usepackage{bm}
\usepackage{marvosym}
\usepackage{graphics}

\usepackage{url}
\usepackage{color}
\usepackage{epstopdf}
\usepackage{subfigure}

\usepackage{breqn}
\usepackage{multirow}
\usepackage{booktabs}
\usepackage{subfigmat}
\usepackage{subfig}
\newtheorem{Def}{Definition}

\newcommand{\tabincell}[2]{\begin{tabular}{@{}#1@{}}#2\end{tabular}}
\usepackage{url}
\usepackage{lipsum}

\begin{document}
%

\title{RiskOracle: A Minute-level Citywide Traffic Accident Forecasting Framework}

\author{
\Large \textbf{Zhengyang Zhou,\textsuperscript{\rm 1} Yang Wang,\textsuperscript{\rm 1,2}\thanks{Prof. Yang Wang is the corresponding author.} Xike Xie,\textsuperscript{\rm 1,2}\thanks{Prof. Xike Xie is the joint corresponding author.} Lianliang Chen,\textsuperscript{\rm 1} Hengchang Liu\textsuperscript{\rm 3}}\\ 
\textsuperscript{\rm 1}School of Computer Science and Technology, University of Science and Technology of China\\ 
\textsuperscript{\rm 2}School of Software Engineering, University of Science and Technology of China\\
\textsuperscript{\rm 3}University of Electronic Science and Technology of China\\
\{zzy0929, cll006\}@mail.ustc.edu.cn, \{angyan,$^*$ xkxie,$^\dag$\}@ustc.edu.cn, liu.heng.chang@gmail.com 
}


\maketitle

\begin{abstract}
Real-time traffic accident forecasting is increasingly important for public safety and urban management (e.g., real-time safe route planning and emergency response deployment). 
Previous works on accident forecasting are often performed on hour levels, utilizing existed neural networks with static region-wise correlations taken into account. However, it is still challenging when the granularity of forecasting step improves as the highly dynamic nature of road network and inherent rareness of accident records in one training sample, which leads to biased results and zero-inflated issue.
In this work, we propose a novel framework RiskOracle, to improve the prediction granularity to minute levels. Specifically, we first transform the zero-risk values in labels to fit the training network. Then, we propose the Differential Time-varying Graph neural network (DTGN) to capture the immediate changes of traffic status and dynamic inter-subregion correlations. Furthermore, we adopt multi-task and region selection schemes to highlight citywide most-likely accident subregions, bridging the gap between biased risk values and sporadic accident distribution. Extensive experiments on two real-world datasets demonstrate the effectiveness and scalability of our RiskOracle framework.
\end{abstract}

\section{Introduction}

Traffic accident forecasting is of great significance for urban safety. For example, with the deployment of Tennessee accident prediction model, the fatality rate of Tennessee has been reduced by 8.16\% in 2016, according to statistics~\cite{Tennessee}. There has been an increasing demand to conduct the accident prediction in a finer granularity, enabling more timely safe route recommendation for travelers and accurate emergency response for emerging applications, such as traffic intelligence and automatic driving.

Regarding the length of accident prediction periods, existing tasks of traffic accident forecasting are classified into two parts, long-term (day-level prediction) and mid-term (hour-level prediction). We summarize all these related works in Table~\ref{tab:relatedwork}. Even though recent studies focus on day-level forecasting~\cite{11,8,Tennessee} by modeling the spatiotemporal heterogeneous data take effects, and~\cite{8} reaches the state-of-the-art performance, it is less meaningful for emergency conditions. 

Mid-term accident forecasting on hour levels can be further classified into classic learning- and deep learning-based methods. Classic learning models include clustering based~\cite{13}, frequent tree based~\cite{12} and Nonnegative Matrix Factorization based~\cite{15} methods. Unfortunately, this type of methods ignores the temporal relations and cannot model the complex nonlinear spatiotemporal interactions. Deep learning-based methods such as~\cite{10} utilize LSTM layers to learn the temporal relations by feeding only historical traffic accident records into the training network, which lacks multi-source real-time traffic inputs to support the forecasting, leading to unsatisfactory prediction performance. There have been works~\cite{6,7,9} on investigating accident patterns through existing deep learning frameworks SDAE, SDCAE and ConvLSTM, respectively, by incorporating real-time human mobilities. However, they all fail to extract the time-varying inter-subregion and intra-subregion correlations in the whole city. 


Although recent advances in deep learning models enable promising results in hour-level accident forecasting, we argue that three important issues are largely overlooked, resulting in poor performance in prediction on minute levels. Firstly, as mentioned in~\cite{9}, when the spatiotemporal resolution of the prediction tasks improves, zero-inflated problems will occur, predicting all results as zeros. Without any strategy to deal with this issue, rare non-zero items in training data disable models to take effects~\cite{wang2018graph}. Secondly, although degrees of static subregion-wise correlations can be learned by Convolution Neural Network (CNN)~\cite{7,9}, time-varying subregion-wise correlations also play a vital role in citywide short-term accident prediction, i.e. two subregions tend to be strongly correlated in the morning and less correlated in the afternoon due to the tidal flows. Thirdly, abnormal changes in traffic status within the same subregion during adjacent time intervals usually induce the occurrence of accidents or other events~\cite{chen2018radar,zheng2015detecting}. Without considering aforementioned spatiotemporal issues, the ability of previous hour-level prediction models would be hindered seriously. 

In this paper, we study the problem of minute-level citywide traffic accident prediction by proposing the three-stage framework RiskOracle, based on Multi-task Differential Time-varying Graph convolution Network (Multi-task DTGN). In the data preprocessing stage, we propose a co-sensing strategy to maximumly infer global traffic status and then a priori knowledge-based data enhancement is designed to tackle zero-inflated issue for short-term predictions. In the training stage, we propose Multi-task DTGN, where time-varying overall affinity explicitly models the short-term dynamic subregions-wise correlations and differential feature generator establishes high-level relationships between the immediate changes of traffic status and accidents. As we know, the accidents and traffic volumes are often distributed imbalanced in the city, thus the multi-task scheme is to address spatial heterogeneities in accident prediction. Then we can obtain a set of discrete most-likely subregions by taking advantages of the learned multi-scale accident distributions in the prediction stage. Experiments on two datasets demonstrate that our framework surpasses state-of-the-art solutions on both 30-minute and 10-minute level prediction tasks.

\begin{table}[]	\footnotesize
	\centering
	\caption{Summarization of traffic accident prediction}

	\begin{tabular}{ccc}
	
		\hline
		\tabincell{c}{Time\\granularity}     & \tabincell{c}{Classic learning \\based method}   & \tabincell{c}{Deep learning \\based method}                  \\   \hline
		Day-level            & \tabincell{c}{\cite{11}\\(Tennessee\\model 2017)
		}      &\tabincell{c}{(Yuan, Zhou and\\ Yang 2018)}                              \\ \hline
		Hour-level           & \tabincell{c}{\cite{13}\\ \cite{12}\\\cite{15}}     & \tabincell{c}{\cite{6}\\ \cite{7}\\ \cite{9} \\ \cite{10}}  \\ \hline
		Minute-level         & -                      & \textbf{Our work}      \\ \hline    
		\label{tab:relatedwork}                         
	\end{tabular}
\end{table}

\section{Preliminaries and Problem Defintion}

In this section, we present the preliminaries and basic definitions, then formally define the problem studied in this paper.

In our work, we find that it leads to unnecessary redundancies, if directly modeling the whole study area as an overall square-shaped region and adopting traditional CNN for spatiotemporal feature extraction, especially for real-time accident forecasting, as the contour of a city is usually irregular. As Figure~\ref{fig:accdistr}(a) shows, we first divide the study area within the road network into $q$ medium-sized rectangular regions (rectangular regions in short). Each rectangular region consists of several small-sized square subregions (subregions in short). There are total $m$ subregions in the study area, and we model the $m$ subregions by the urban graph. 

{\em \begin{Def}[\textbf{Urban Graph.}]
The study area can be defined as an undirected graph, called urban graph $G(\mathcal{V},\mathcal{E})$. Here, the vertex set $\mathcal{V} = \{ {v_1},{v_2}, \cdots ,{v_m}\}$, where $v_i$ denotes the $i$-th square-shaped urban subregion. Given two vertexes ${v_i},{v_j} \in \mathcal{V}$, the edge ${e_{ij}} \in \mathcal{E}$ within these two vertexes indicates the connectedness between these two subregions, where
\begin{equation}
{e_{ij}} = \left\{ {\begin{array}{*{20}{c}}
1&\begin{array}{l}
{\rm{iff}}\;{\rm{the}}\;{\rm{traffic}}\;{\rm{elements}}\;{\rm{within}}\;{\rm{two}}\\
\,{\rm{subregions}}\;{\rm{have}}\;{\rm{strong}}\;{\rm{correlations}}
\end{array}\\
{}&{}\\
0&{{\rm{otherwise}}}
\end{array}} \right.
\end{equation}
\end{Def}}
{\rm Note that in this paper, the traffic elements of a vertex consist of two aspects, static road network features and dynamic traffic features. And we keep the $\rho$ connectedness of the whole urban graph to control sparsity of affinity matrix $\mathcal{A}_{s}$ and $\mathcal{A}_{o}^{\Delta t}$ (introduced in the next section), then the corresponding nonzero items in affinity matrix refer to the subregions with strong correlations.}

{\rm The dynamic traffic features of subregion $v_i$ in a specific time interval $\Delta t$ can be modeled by three parts, (a) the intensity of human activities, represented by traffic volume $TV_{v_i}(\Delta t)$; (b) the traffic conditions, represented by the average traffic speed $a_{v_i}(\Delta t)$; and (c) the level of traffic accident risks $r_{v_i} (\Delta t)$. The formal definition of dynamic traffic features is as follows.}

{\em \begin{Def}[\textbf{Static Road Network Features.}]
For an urban subregion ${v_i} \in \mathcal{V}$, the static features of road networks within the subregion, cover the statistical spatial attributes of the numbers of road lanes, road types, road segment lengths and widths, snow removal priorities and the numbers of overhead electronic signs, for all road segments inside, can be denoted as a fixed length vector $s_i$. The static road network features of the entire urban domain can be formulated as $\mathcal{S} = \{ {s_1},{s_2}, \cdots ,{s_m}\} $.
\end{Def}}

{\em \begin{Def}[\textbf{Dynamic Traffic Features.}]
For ${v_i} \in \mathcal{V}$, the dynamic traffic features of $v_i$ within a given time interval $\Delta t$ can be formulated as $f_{v_i}({\Delta t})=\{T{V_{{v_i}}}(\Delta t),{a_{{v_i}}}(\Delta t),{r_{{v_i}}}(\Delta t)\}$. $r_{v_i} (\Delta t)$ is the summation of the number of accidents weighted by the corresponding severity levels \footnote{We define three accident risk types: minor accidents, injured accidents, and fatal accidents~\cite{6}. We assign weights 1, 2, and 3 to the three types, respectively.}. In particular, ${r_{{v_i}}}(\Delta t) = \sum\limits_{j = 1}^3 j *\tau _{{v_i}}^{\Delta t}(j)$, where $j$ indicates the type of accident severity, $\tau _{{v_i}}^{\Delta t}(j)$ denotes the number of accidents of type $j$. So the accident risk distributions and the dynamic traffic features of the entire urban domain within $\Delta t$ can be represented by $\mathcal{R}(\Delta t) = \{ {r_{{v_1}}}(\Delta t),{r_{{v_2}}}(\Delta t), \cdots ,{r_{{v_m}}}(\Delta t)\}$ and $\mathcal{F}(\Delta t) = \{ {f_{{v_1}}}(\Delta t),{f_{{v_2}}}(\Delta t), \cdots ,{f_{{v_m}}}(\Delta t)\} $ respectively.
\end{Def}}


{\em \begin{Def}[\textbf{Traffic Accident Prediction.}]
Given static road network features $\mathcal{S}$ and the historical dynamic traffic features $\mathcal{F}(\Delta t)\left ({\Delta t = 1,2, \cdots T}) \right.$, our purpose is to predict the distribution of the citywide traffic accident risks $\mathcal{R}(T + 1)$ and select high-risk subregions $\mathcal{V}_{\mathit{acc}}(T+1)$, for the future time interval $T+1$.
\end{Def}} 
\section{Minute-level Real-time Traffic Accident Forecasting}

In this section, we first show the overview of our proposed framework RiskOracle, and then elaborate on each stage.

\subsection{Framework Overview}

\begin{figure}[!]
   \centering
   \includegraphics[width=.95\columnwidth]{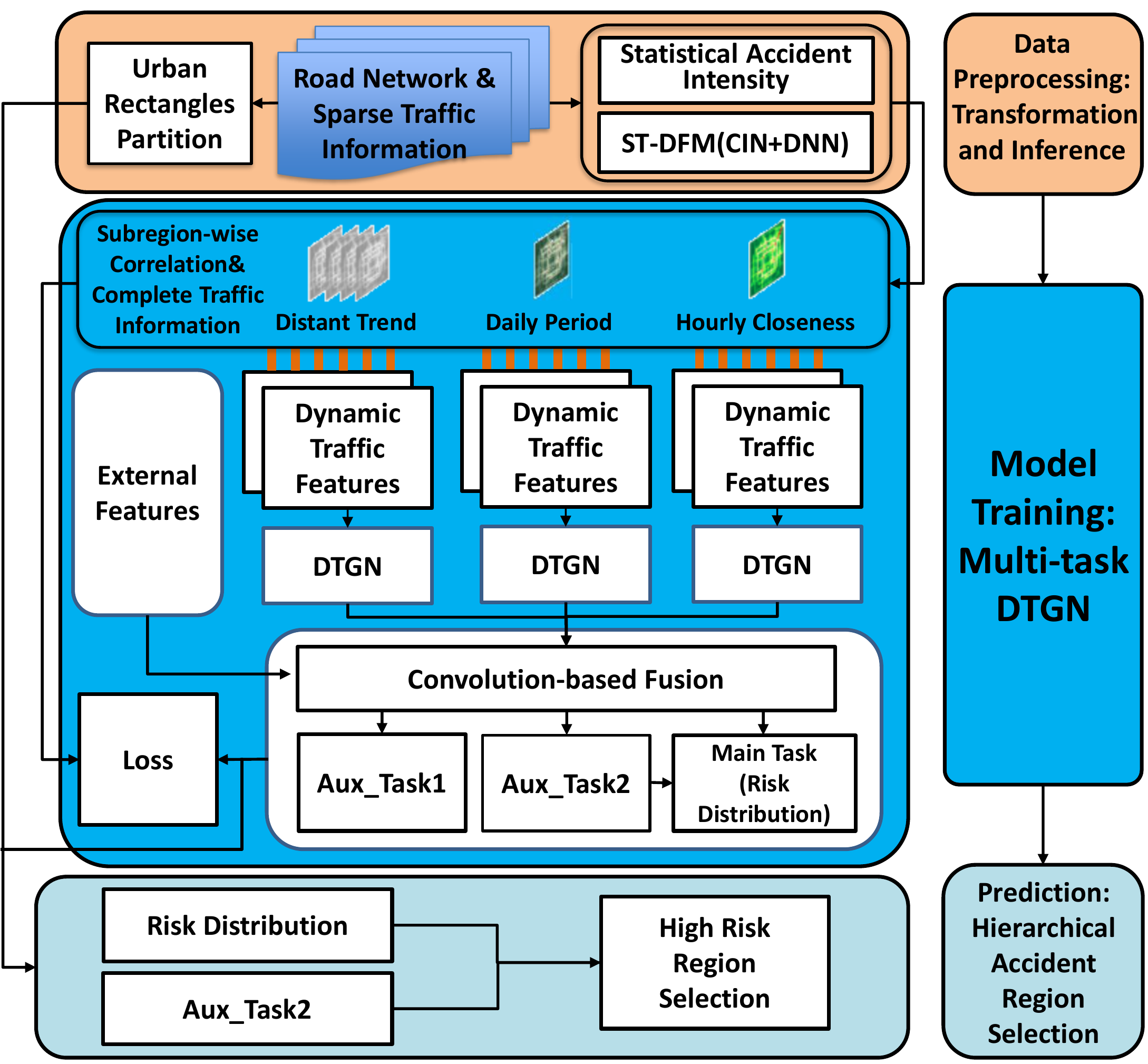}
   \caption{Framework Overview of RiskOracle}
   \label{fig:so}

\end{figure}

As illustrated in Figure~\ref{fig:so}, our proposed framework RiskOracle includes three stages, Data preprocessing stage, Model training stage and Prediction stage. 

\subsection{Data Preprocessing}
\subsubsection{Addressing Spatial Heterogeneities in Accident Prediction}
High-risk values tend to bias to urban areas due to most accidents and traffic volumes are covered downtown, leading to a serious spatial imbalance in risks and ignoring the relatively high-risk regions in rural areas. To perform citywide predictions, it is necessary to select most-likely accident regions and address the spatial heterogeneities. 
Thus, as illustrated in Figure~\ref{fig:accdistr}(a), the subregions are organized hierarchically in our work and they are responsible for collecting fine- and coarse-grained accident distributions, respectively. Then subregions in each medium-sized rectangular region will be further highlighted separately. The multi-scale distributions can be considered as the hierarchical accident distributions.

\subsubsection{Overcoming Zero-inflated Issue}
\begin{figure}[!ht]	
	\centering
	\includegraphics[width=.95\columnwidth]{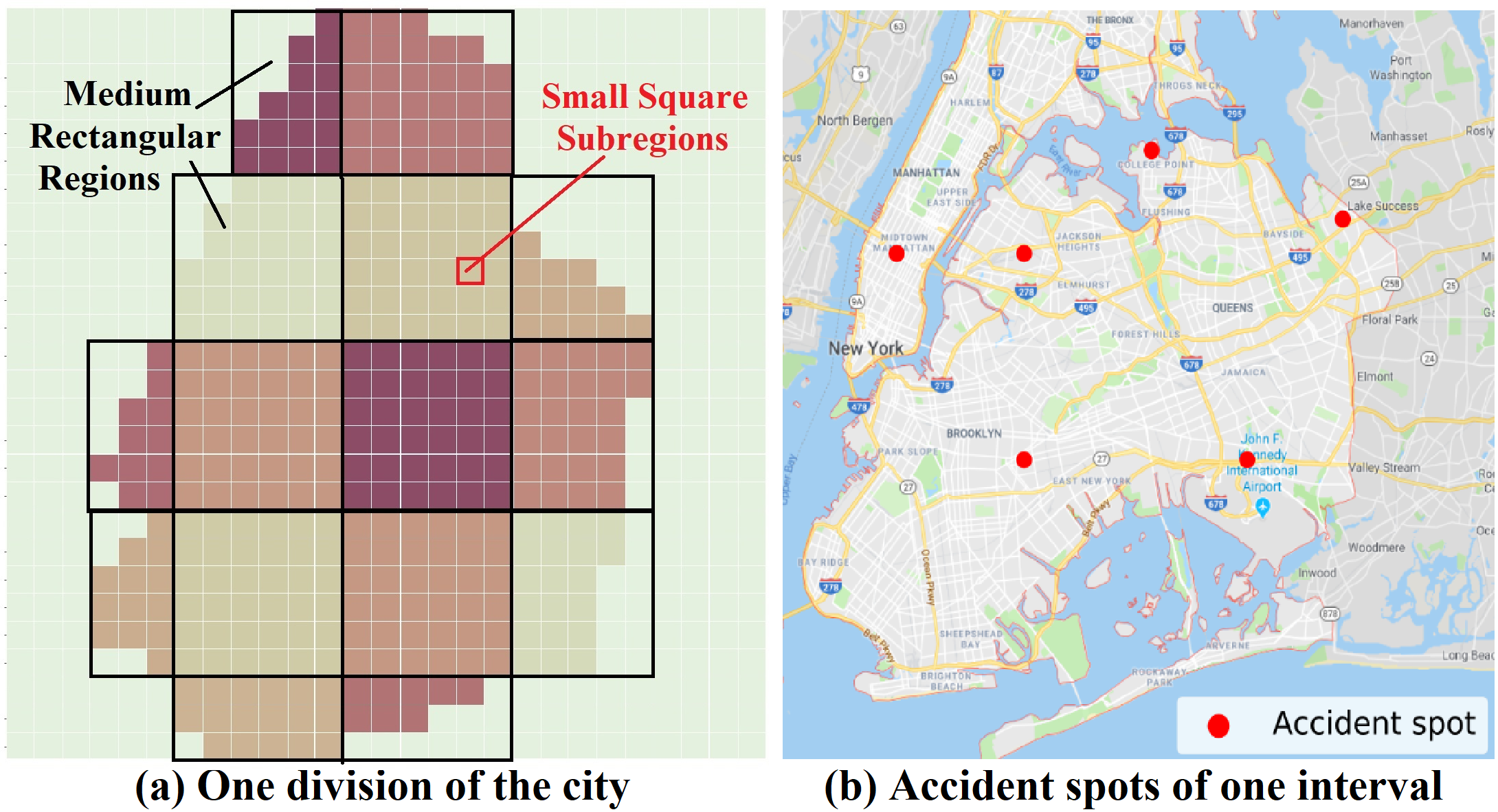}
	\caption{An example of NYC}
	\label{fig:accdistr}
\end{figure}

Deep Neural Networsks (DNNs) suffer from zero-inflated issues and predict invalid results if the nonzero items in training labels are extremely rare~\cite{wang2018graph,9}. There only exist 6 accidents in the whole study area during a selected 10-minute interval in New York City (NYC) as Figure~\ref{fig:accdistr}(b) shows, demonstrating the inherent rareness of short-term accidents. To overcome this issue in real-time accident prediction, we devise a priori knowledge-based data enhancement (PKDE) strategy to discriminate the risk values in labels of training dataset. Specifically, for interval $\Delta t$, we transform zero items in $\mathcal{R}(\Delta t)$ to negative values. The transformation is done in two phases: a) the zero value is transformed into accident risk indicator by Equation (2); b) the value of indicator is transformed into statistical accident intensity by Equation (3). Given subregion $v_i$, we can calculate its accident risk indicator ${\varepsilon _{{v_i}}}$
by
\begin{equation}
{\varepsilon _{{v_i}}} = \frac{1}{{{\mathit{N_{week}}}}}\sum\limits_{j = 1}^{{\mathit{N_{week}}}} {\frac{{{r_{{v_i}}}(j)}}{{\sum\limits_{k = 1}^m {{r_{{v_k}}}(j)} }}} 
\end{equation}
where $\mathit{N_{week}}$ is the total number of weeks in the training dataset, and ${{r_{{v_i}}}(j)}$ indicates the total risk value of region $v_i$ during all time intervals in the $j$-th week. Then, we can calculate the statistical accident intensity of region $v_i$ by 
\begin{equation}
{\pi _{{v_i}}} = {b_1}{\log _2}{\varepsilon _{{v_i}}} + {b_2}
\end{equation}
where $b_1$ and $b_2$ are the coefficients to maintain symmetry between the range of the absolute value of $\pi_{v_i}$ and the range of true risk values. With the logarithm transformation within 0 and 1, we can easily make transformed data discriminating and suitable for training networks. The transformation is implemented in such a way: 1) the accident intensity value of a zero-item subregion is negative and thus smaller than the value of nonzero-item subregion, reflecting the fact that a zero-item subregion is with lower accident risk; 2) the subregion with lower accident risk indicator has a lower accident probability, preserving the ranks of actual accidient risks. 

\subsubsection{Complementing Sparse Sensing Data}
Real-time traffic information is usually collected insufficiently~\cite{wang2018real} for accident prediction and the dynamic traffic information tends to have interactive effects with the static spatial road network structures~\cite{18,12}. Thus, we propose a co-sensing strategy by modifying xDeepFM~\cite{lian2018xdeepfm} as SpatioTemporal Deep Factorization Machine (ST-DFM) by taking advantages of the interaction operations of FM. 

We first extract the road network similarities and connections between subregions by static affinity matrix $\mathcal{A}_{s}$ where the item $\alpha_{s}(i,j)$ in $\mathcal{A}_{s}$ denotes static affinity within subregion $v_i$ and $v_j$ and the affinity can be calculated by
\begin{equation}
{\alpha_{s}}(i,j) = \left\{ {\begin{array}{*{20}{c}}
	1&\begin{array}{l}
	{\rm{if}}\;{\rm{subregion}}\;{v_i}\;{\rm{and}}\;\\{v_j}\;
	\;\;\;\;{\rm{are}}\;{\rm{adjacent}}
	\end{array}\\
	{}&{}\\
	{{e^{ - \mathit{JS}({s_i}\left\| {{s_j}} \right.)}}}&{{\rm{otherwise}}}
	\end{array}} \right.
\end{equation}
Here, the $\mathit{JS}$ function is the Jensen-Shannon divergence~\cite{lin1991divergence}:
\begin{equation}
JS({s_i}\left\| {{s_j}} \right.) = \frac{1}{2}\sum\limits_k {\left( {\begin{array}{*{20}{c}}
		{{s_i}(k)\log \frac{{2{s_i}(k)}}{{{s_i}(k) + {s_j}(k)}} + }\\
		{{s_j}(k)\log \frac{{2{s_j}(k)}}{{{s_i}(k) + {s_j}(k)}}}
		\end{array}} \right)} 
\end{equation}
The same as xDeepFM, ST-DFM contains Compressed Interaction Network (CIN) module and DNN module. Three spatiotemporal fields i.e. static spatial features, dynamic traffic features\footnote{\footnotesize For the dynamic traffic feature field in one subregion $v_i$, we first select the most proximal subregions with $v_i$ by the static affinity matrix, the available dynamic traffic information within these subregions will constitute of the dynamic traffic features in $v_i$.} and timestamps are embedded in ST-DFM. Then, ST-DFM learns the interactive relationships between different spatiotemporal features in vector-wise level with the CIN module and the high-level representation of features with the DNN module, and finally obtains high-level feature combinations. We infer speed values by feeding traffic volumes at the corresponding subregion into ST-DFM and vice versa. Then traffic information can thus be maximumly inferred to obtain global traffic status by training the data within the intersections of two real-time traffic datasets.

\subsection{Multi-task DTGN for Accident Risk Prediction}
\subsubsection{SpatioTemporal DTGN}
The accidents and congestions tend to be interacted and propagated in the road network, especially on holidays or in rush hours. Due to the potential of GCN in modeling non-Euclidean subregion-wise propagations and correlations~\cite{18}, we hereby propose DTGN. We modify GCN by incorporating time-varying overall affinity and differential feature generator to tackle the challenges in minute-level accident prediction. 

\textbf{Time-varying overall affinity matrix with dynamic traffic features involved. } It has been demonstrated strong time-varying correlations between traffic conditions of different urban subregions~\cite{26,wang2014data}. Also, there exist strong spatiotemporal correlations between traffic accidents and urban traffic conditions~\cite{chen2018radar}. Therefore, for our minute-level accident prediction, it is indispensable to capture the inter-subregion time-varying traffic correlations of a specific time interval $\Delta t$ by an overall affinity matrix $\mathcal{A}_{o}^{\Delta t}$. The item $\alpha_{o}^{\Delta t}(i,j)$ in $\mathcal{A}_{o}^{\Delta t}$ denotes the dynamic overall affinity within subregions $v_i$ and $v_j$:
\begin{equation}
\alpha _{o}^{\Delta t}(i,j) = {e^{ - \mathit{JS}(s_i^*\left\| {s_j^*} \right.)}} + \gamma *{e^{ - \mathit{JS}(C_i^{\Delta t}\left\| {C_j^{\Delta t}} \right.)}}
\end{equation}
$C_i^{\Delta t}$ includes the traffic volume $T{V_{{v_i}}}(\Delta t)$ and average speed ${a_{{v_i}}}(\Delta t)$ of subregion $v_i$ within the same time interval $\Delta t$ in each day of last week. Notice that we modify the weights of static spatial attributes of subregions based on their different effects on accidents with an attention-based scheme~\cite{bahdanau2014neural}. Also, the accident-based static features of subregion $v_i$ can be denoted as $s_i^*$. Further, a weighted factor $\gamma$ is used to adjust the proportion that the dynamic traffic condition affinity accounts for the overall affinity matrix. With such overall affinity, distant subregions but have potential accident-related correlations due to traffic characteristics can also be connected dynamically. To perform GCN in spectral domain, we need to calculate the Laplacian matrix ${L^{\Delta t}}$~\cite{23} with $\mathcal{A}_{o}^{\Delta t}$, which can be seen as the graph adjacence martix. First, we derive $\mathcal{B}^{\Delta t}$: 
\begin{equation}
\mathcal{B}^{\Delta t}  = \mathcal{A}_o^{\Delta t} + {I_m}
\end{equation}
where $I_m$ is the identity matrix of $m \times m$. Second, we calculate ${\Phi ^{\Delta t}}$ by
\begin{equation}
{\Phi ^{\Delta t}} = \left[ {\begin{array}{*{20}{c}}
{{\varphi _{11}}}&0& \cdots &0\\
0&{{\varphi _{22}}}& \cdots &0\\
 \vdots & \vdots & \ddots & \vdots \\
0&0& \cdots &{{\varphi _{mm}}}
\end{array}} \right]
\end{equation}
where ${\varphi _{ii}} = \sum\limits_{j = 1}^m {{b_{ij}}} $ and $b_{ij}$ is the element in matrix $\mathcal{B}^{\Delta t}$. Then, we can obtain Laplacian matrix of $\Delta t$ by
\begin{equation}
{L^{\Delta t}} = {\left( {{\Phi ^{\Delta t}}} \right)^{ - \frac{1}{2}}}{{\cal B}^{\Delta t}}{\left( {{\Phi ^{\Delta t}}} \right)^{ - \frac{1}{2}}}
\end{equation}

\begin{figure}[!]

  \includegraphics[width=.95\columnwidth]{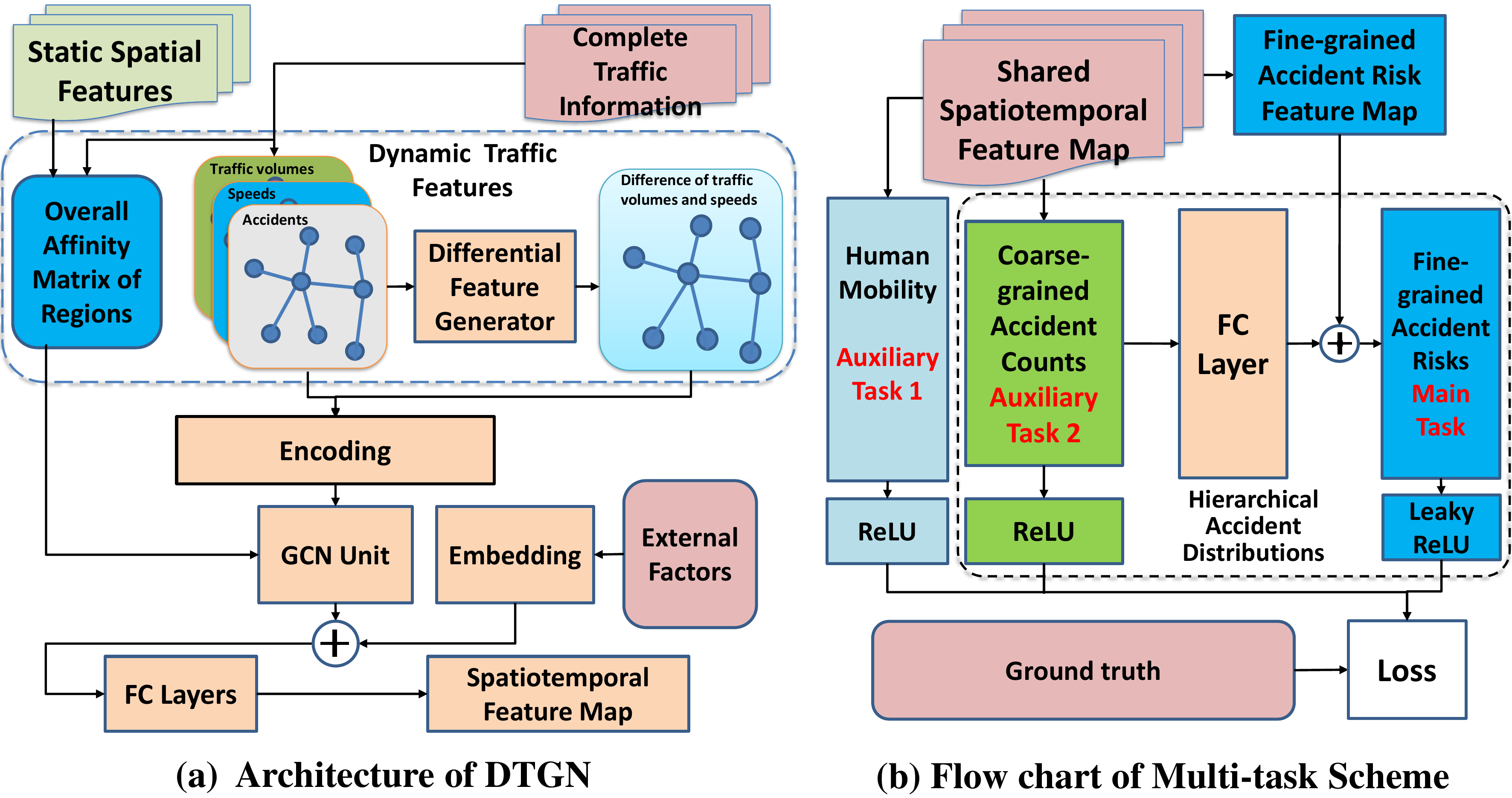}

   \caption{Details of Multi-task DTGN}
   \label{fig:adtgn}

\end{figure}

\textbf{Differential GCN for extracting spatiotemporal features. } It has been generally accepted that the task of accident or event prediction is more relevant to abnormal variations of urban traffic conditions, compared with regular traffic conditions~\cite{chen2018radar,zheng2015detecting}. To this end, we introduce a differential feature generator to calculate differential images within adjacent time intervals. By feeding the differential dynamic traffic features into GCN, the propagations and interactions of abnormal changes in traffic can be modeled and the high-level correlations between the immediate traffic status variations and accidents are learned, especially benefiting minute-level accident forecasting. Given $\Delta t$, the differential vector ${\overrightarrow \Theta  ^{\Delta t}}$ can be computed by
\begin{equation}
{\overrightarrow \Theta  ^{\Delta t}} = \mathcal{D}(\Delta t) - \mathcal{D}(\Delta t - 1)\;
\end{equation}
where $\mathcal{D}(\Delta t) = \left\{ {{d_{{v_1}}}(\Delta t),{d_{{v_2}}}(\Delta t), \cdots ,{d_{{v_m}}}(\Delta t)} \right\}$ and ${d_{{v_i}}}(\Delta t) = \{T{V_{{v_i}}}(\Delta t),{a_{{v_i}}}(\Delta t)\}$. 
For all subregions in $\Delta t$, by combining their dynamic traffic features and the corresponding differential vectors, we generate a united feature tuple $\mathcal{U}(\Delta t) = \left\{ {\mathcal{F}(\Delta t),{{\overrightarrow \Theta  }^{\Delta t}}} \right\}$. As described in~\cite{2}, urban traffic has obvious characteristics of three temporal perspectives, hourly closeness, daily periodicity and distant trend. To this end, when given $\Delta t$, we select $\kappa$ united feature tuples~\footnote{According to the settings in~\cite{2}, we set the value of $\kappa$ as 3.} for each temporal perspective as the inputs of DTGNs. 
Specifically, we select the last $\kappa$ intervals of $\Delta t$, the same interval as $\Delta t$ of the last continuous $\kappa$ days for hourly closeness and daily periodicity perspective. And for distant trend, we first select $\kappa$ previous days at the frequency of every 10 days and for each of $\kappa$ selected days, we extract the same interval as $\Delta t$. As illustrated in Figure~\ref{fig:so}, we then feed the united feature tuple sets of all three temporal perspectives into DTGNs independently. The detailed architecture of one DTGN is demonstrated in Figure~\ref{fig:adtgn}(a). For one specific temporal perspective, we denote the corresponding united feature tuple set as $\mathbb{U}_*^{\Delta t}$. We feed  $\mathbb{U}_*^{\Delta t}$ into a fully-connected (FC) network to encode all features into a lower-dimensional feature set, and then feed it into a GCN. The GCN works recursively,
\begin{equation}
{\mathcal{H}^{n + 1}} = {\mathop{\rm Leaky\_ReLU}\nolimits} ({L^*}{\mathcal{H}^{n }}{\mathcal{W}^n})\;{\rm{where}}\;{\mathcal{H}^0} = \mathbb{U}_*^{\Delta t}
\end{equation}
Here $\mathcal{H}^{n }$ indicates the $n$-layer graph convolution, $\mathcal{W}^n$ denotes the weights of the $n$-layer graph convolution kernel. Notice here, given one temporal perspective, we take the mean of the $L^{\Delta t}$ matrices of all selected time intervals as $L^*$. We use a Batch Normalization between every 2 GCN layers to avoid gradient explosion. Considering negative values in the dataset we transformed, we select $\mathop{\rm Leaky\_ReLU}$ as the activation function. In addition, real-time dynamic external factors, i.e. timestamps and meteorological data, are embedded into a vector of fixed length consecutively, and then fused with the output of each GCN unit. For three temporal perspectives, we denote the output feature maps of DTGN as $\mathcal{O}_{hc}^{\Delta t}$, $\mathcal{O}_{dp}^{\Delta t}$ and $\mathcal{O}_{dt}^{\Delta t}$ respectively. 

\subsubsection{Multi-task Learning for Accident Risk Prediction}
In this subsection, we design the multi-task scheme, not only to enhance the deep representation, also to learn hierarchical accident distributions and provide instructions for most-likely accident region selection. For forecasting accident risks of subregions, we first take the distribution of accident risks as the main task. Considering the prominent correlations between traffic accidents and the intensity of human activities, we take regional traffic volume prediction as the first auxiliary task to enhance the representation. To provide instructional information for the hierarchical accident region selection, we take the total numbers of accidents within different rectangular regions as the second auxiliary task. 

Specifically, we feed the output feature maps of DTGN including $\mathcal{O}_{hc}^{\Delta t}$, $\mathcal{O}_{dp}^{\Delta t}$ and $\mathcal{O}_{dt}^{\Delta t}$ into a convolution-based fusion module as Figure~\ref{fig:so} shows, then perform the multi-task learning. We visualize the flow chart of our multi-task scheme in Figure~\ref{fig:adtgn}(b). First, we generate the predicted risk distribution feature map $\mathcal{O}_{risk}^{\Delta t}$ and the citywide traffic volumes $\mathcal{O}_{vol}^{\Delta t}$ as follows, the reasons why we choose Leaky\_ReLU in the main task due to the risks in labels are partly transformed into negative values and other tasks remain nonnegative.
\begin{equation}
\mathcal{O}_{\mathit{risk}}^{\Delta t} = {\rm{Leaky\_ReLU(}}{{\rm{\mathcal{W}}}_{\mathit{risk}}^{\Delta t}}*[\mathcal{O}_{hc}^{\Delta t},\mathcal{O}_{dp}^{\Delta t},\mathcal{O}_{dt}^{\Delta t}]{\rm{)}}
\end{equation}
\begin{equation}
\mathcal{O}_{\mathit{vol}}^{\Delta t} = {\rm{ReLU(}}{{\rm{\mathcal{W}}}_{\mathit{vol}}^{\Delta t}}*[\mathcal{O}_{hc}^{\Delta t},\mathcal{O}_{dp}^{\Delta t},\mathcal{O}_{dt}^{\Delta t}]{\rm{)}}
\end{equation}
With the additional fully-connected layer, we learn the total number of accidents within each rectangular region by:
\begin{small}
\begin{equation}
{\cal O}_{\mathit{count}}^{\Delta t} ={\rm{ReLU}} \left( {\mathcal{W}_{fc}^{\Delta t}}*({{\rm{\mathcal{W}}}_{\mathit{count}}^{\Delta t}*[{\cal O}_{hc}^{\Delta t},{\cal O}_{dp}^{\Delta t},{\cal O}_{dt}^{\Delta t}])}\right)
\end{equation}
\end{small}
Here, $\mathcal{W}_{\mathit{risk}}^{\Delta t}$, $\mathcal{W}_{\mathit{vol}}^{\Delta t}$ and $\mathcal{W}_{\mathit{count}}^{\Delta t}$ denote the fusion weights of accident risk, human activity intensity and the numbers of accidents within different medium rectangular regions in time interval $\Delta t$, respectively.  $\mathcal{W}_{\mathit{fc}}^{\Delta t}$ is the fusion weights of the fully-connected network in $\Delta t$. And ${\cal O}_{\mathit{count}}^{\Delta t}$ can be viewed as the coarse-grained accident distribution, it will be fed into another fully-connected layer to map to same shape as $\mathcal{O}_{\mathit{risk}}^{\Delta t}$, and then intergrated with the output of accident risk distribution feature map $\mathcal{O}_{\mathit{risk}}^{\Delta t}$, compelling both tasks to learn the relationship between multiple accident distributions adequately. Then $\mathcal{O}_{\mathit{risk}}^{\Delta t}$ can be updated by
\begin{small}
\begin{equation}
{\cal O}_{\mathit{risk*}}^{\Delta t}= {\rm{Leaky\_ReLU}} ({\mathcal{W}_{\mathit{fc*}}}*{\cal O}_{\mathit{count}}^{\Delta t} + \mathcal{O}_{\mathit{risk}}^{\Delta t})
\end{equation}
\end{small}
where ${\cal O}_{\mathit{risk*}}^{\Delta t}$ is the final output of main task and $\mathcal{W}_{\mathit{fc*}}$ denoting the weights of the fully-connected layer.
So we have the total loss of this multi-task learning framework as
\begin{equation}
{\rm Loss}(\theta ) = ms{e_{risk*}} + {\lambda _1}*ms{e_{vol}} + {\lambda _2}*ms{e_{count}} + {\lambda _3}*{L_2}
\end{equation}
where $ms{e_{risk*}}$, $ms{e_{vol}}$ and $ms{e_{count}}$ are the loss of the main task and two auxiliary tasks respectively. We use L2 regularization to avoid the overfitting issue, and use $\lambda _1$, $\lambda _2$, $\lambda _3$ as the hyper-parameters of the loss function.

\subsection{Hierarchical Most-likely Accident Region Selection}
In a specific city, there often exist imbalanced coverage of accidents and traffic volumes in rural and urban areas, inducing the issue of spatial heterogeneities. Thus, it is illogical to cut off high accident risk with a unified risk threshold for selecting most-likely accident subregions. We then propose a hierarchical most-likely accident region selection (HARS) strategy based on the hierarchical accident distributions learned in the multi-task scheme. 

For each rectangular region $i$, we select $k_i(i=1,2,...,q)$ subregions with the highest risks and the parameter $k_i$ equals to the corresponding element in ${\cal O}_{\mathit{count}}^{\Delta t}$ learned by the second auxiliary task. In consequence, we obtain a set of most-likely accident regions. Also, the learned $k_i$ reduces the overpredicted regions and keeps the model conform to the changes of time and weather with external factors involved.
\section{Empirical Studies}
In this section, we conduct extensive empirical studies to evaluate our minute-level prediction framework by setting the temporal intervals as 30 minutes and 10 minutes. 

\subsection{Data Description}
We conduct experiments on two real-world datasets: NYC Opendata and Suzhou Industrial Park (SIP) dataset. For NYC dataset, due to the lack of real-time traffic volumes, here we utilize the taxi trip volumes in each subregion as the indicator of human mobilities. For SIP dataset, it contains traffic flows and speeds. We integrate it with another traffic accident dataset collected from Microblog, Sina, a social media platform. The statistics are shown in Table~\ref{tab:datasets}. More details are available on the website\footnote{https://github.com/zzyy0929/AAAI2020-RiskOracle/.}. 

\subsection{Implementation Details}
For experiments, we select 60\%, 30\% and 10\% of dataset for training, evaluation and validation, respectively. We generate the subregion set $\mathcal{V}$ by partitioning the city map into small subregions with equal size referring to common settings~\cite{18} and practices. We stack 9 GCN layers with 384 filters in each layer. The weights of the loss function are set as ${\lambda _1} = 0.8$, ${\lambda _2} = 1$, ${\lambda _3} ={1{\rm{e}} - 4} $. The multi-task DTGN is trained with back propagation and Adam method~\cite{kingma2014adam}.

During training period, dynamic traffic data and affinity matrices are aggregated into 3 groups and two-scale accident distributions are fed into Multi-task DTGN. For testing, we fetch the needed data and pass it through the model, most-likely accident subregions are derived with main and second auxiliary task. The high-risk subregions are highlighted and compared with real-world accident records during the same spatiotemporal scope.

\begin{table}[]\footnotesize
	\centering
	\caption{Datasets statistics}
	
	\begin{tabular}{ccccc}
		\toprule
		City & Dataset\tablefootnote{It refers to different types of traffic-related records in the city.}               & Time Span                          & {\tabincell{c}{\# of \\ Regions}}         &  {\tabincell{c}{\# of \\ Records}} \\ \hline
		\multirow{6}{*}{NYC}  & Accidents      & \multirow{4}{*}{\tabincell{c}{01/01/2017-\\ 05/31/2017}}  & \multirow{6}{*}{354} & 254k       \\
		& Taxi Trips  &                                       &                      & 48,496k    \\
		& Speed Values      &                                       &                      & 125k       \\
		& Weathers     &                                       &                      & 604           \\\cline{2-3}
		& Demographics   & \multirow{2}{*}{\tabincell{c}{Investigated \\ in 2016} } &                      & 195           \\
		& Road Network &                                       &                      & 102k       \\ \hline
		\multirow{4}{*}{SIP}  & Accidents   & \multirow{4}{*}{\tabincell{c}{01/01/2017-\\ 03/31/2017}} & \multirow{4}{*}{108} & 183           \\
		& Traffic Flows  &                                       &                      & 1,399k     \\
		& Speed Values          &                                       &                      & 311k       \\
		& Weathers
		       &                                       &                      & 180          \\ \hline
		\bottomrule
		\label{tab:datasets}
	\end{tabular}

\end{table}

\subsection{Evaluation Metrics}
We evaluate our proposed RiskOracle from two perspectives, regression perspective and classification perspective. (1) Regression perspective: Mean Square Error (MSE) of predicted risks. (2) Spatial classification perspective: a) Accuracy of top $M$ (Acc@$M$)~\cite{liao2018predicting}, which is widely applied in spatiotemporal ranking tasks, indicates the percentage of accurate predictions in subregions within $M$ highest risks. $M$ equals 20 and 6 for 30-minute and 10-minute evaluation in NYC dataset according to the statistics (NYC Accident Records 2017). And similarly, in SIP dataset, $M$ equals 5. b) Acc@$K$, where $K$ is the summation of $k_i$ learned by the second auxiliary task. Note that Acc1 denotes the accuracy during hours with a high frequency of accidents, i.e. 7:00 a.m.-9:00 a.m. and 12:00 p.m.-4:00 p.m.

\subsection{Baselines}

Five baselines are as follows:
\textbf{(1) ARIMA}, a classic machine learning algorithm, for understanding and predicting future values, especially for time-series predictions; \textbf{(2) Hetero-ConvLSTM}, the state-of-the-art deep learning framework for traffic accident prediction\footnote{\footnotesize We adjust the hyper-parameters to reach its best performance at 4 blocks with 16 filters, and a size of 12$\times$12 moving window with step=6.}~\cite{8}; \textbf{(3) ST-ResNet}, proposed in \cite{2} for predicting traffic flows; \textbf{(4) SDAE}, proposed in \cite{6} for real-time risk prediction, by incorporating human mobilities; \textbf{(5) SDCAE}, the latest method for citywide hour-level accident risk prediction proposed in \cite{7}.

\subsection{Evaluation Results and Analysis}

\subsubsection{Comparison Performances}
Table~\ref{tab:NYC30-10} illustrates the performance comparisons on NYC and SIP datasets with 30-minute and 10-minute intervals settings. Encouragingly, our framework RiskOracle achieves the highest accuracy and outperforms baselines on almost all metrics. With HARS, our model addresses the spatial heterogeneities and overprediction issue in accident prediction by highlighting $k_i$ subregions in the $i$-th rectangular region. Especially on NYC dataset, our RiskOracle improves the accuracy by 22.49\% compared with the best baseline on Acc@20. In consequence, RiskOracle is more sensitive to accidents and extensible for sparse sensing data as well as short-term sporadic spatiotemporal forecasting. Additionally, it shows that our model performs better during high-risk hours, which is desired by real applications in accident forecasting. The reasons why the performance on NYC can be better than SIP dataset may be the incompletion of accident labels in SIP. We will report Acc@$K$ later in ablation studies. 

Overall, as the temporal granularity becomes finer, the performances of our framework decrease slightly while baselines decrease sharply as they trap into zero-inflated issue, which demonstrates the effectiveness and scalability of our proposal for short-term accident prediction. The improvements on both two datasets verify the robustness and generality of our proposed RiskOracle even when the dataset in real applications includes rare accident records. 

\begin{table*}[]\footnotesize
	\centering
	\caption{Performance comparisons on NYC and SIP datasets}
	\label{tab:NYC30-10}
	\begin{tabular}{c|c|ccc|ccc}
		\toprule
		\hline
		 & & \multicolumn{3}{c|} {30-minute Interval}     &\multicolumn{3}{c} {10-minute Interval}     \\ \hline
		\multirow{8}{*}{NYC}& Models & MSE & Acc@20 & Acc1@20 & MSE & Acc@6 & Acc1@6  \\ \hline		
		& ARIMA              & 0.6801        & 14.23\%          & 20.26\% &  0.2380    & 8.62\%   & 10.05\%            \\ 
		& Hetero-ConvLSTM    & 0.1129        & 48.04\%          & 58.01\% &  0.0185 & 24.53\% & 42.01\%            \\
		& ST-ResNet & 0.0627 & 31.06\% & 40.93\%         & 0.0162 & 10.02\% & 27.50\%    \\
		& SDAE & 0.2414 & 12.08\% & 27.73\%              & 0.0435 & 8.33\% & 12.74\%     \\
		& SDCAE & 0.2209 & 14.79\% & 22.64\%                 & 0.0076 & 13.48\% & 31.48\% \\
		& \textbf{RiskOracle (Ours)} & \textbf{0.1085} & \textbf{70.53\%} & \textbf{72.91\%} &\textbf{0.0452} & \textbf{45.18\%} & \textbf{69.22\%}  \\ \hline
		\multirow{8}{*}{SIP} & Models & MSE\tablefootnote{\footnotesize Here we report the mse of accidents.} & Acc@5  & Acc1@5  & MSE & Acc@5 & Acc1@5 \\ \hline
		& ARIMA  & -      & 19.68\%  & 18.42\%     & -                & 23.68\% & 28.62\% \\
		& Hetero-ConvLSTM & 3.392     & 28.92\% & 42.37\%  & 3.980  & 31.42\% & 48.57\%  \\ 
		& ST-ResNet    & 3.459     & 60.73\%  & 62.50\%  & 3.180             & 41.78\% & 43.50\%   \\
		& SDAE         & 3.322     & 60.26\%  & 36.72\%  & 3.312            & 12.88\% & 20.83\% \\ 
		& SDCAE        & 3.210     & 58.68\%  & 67.50\%  & 3.455             & 26.31\% & 37.50\%   \\
		& \textbf{RiskOracle(Ours)}     & \textbf{3.270}           & \textbf{63.15\%}  & \textbf{65.24\%}   & \textbf{3.029}           & \textbf{46.30\%}  & \textbf{48.91\%}  \\
		\hline  
		\bottomrule
		
	\end{tabular}
\end{table*}

\subsubsection{Evaluations on Acc@$K$ and Ablation Studies}
In Table~\ref{tab:Ablation-NYC}, we report the results on Acc@$K$ which we propose in our paper. We record $K$ in each interval, which is the summation of $k_i$ learned by our framework for fair comparisons. It is reasonable that results of Acc@20 and Acc@6 are slightly higher than Acc@$K$ because the uniform threshold cannot adapt to the real-time conditions and tend to overpredict the accidents. In contrast, our framework has the flexibility to potentially approximate the number of accidents in each rectangular region with the multi-scale accident distribution forecasting. As observed, our framework can outperform other baselines and achieve an acceptable level of accuracy on Acc@$K$ when compared to results in Table~\ref{tab:NYC30-10}, verifying the effectiveness of our hierarchical accident selection mechanism in the task.

Further, to investigate how each component contributes to high-quality results, we perform an ablation study to tease apart which components of RiskOracle are most important for its success. The prediction performances of ablative variants of RiskOracle are shown in Table~\ref{tab:Ablation-NYC} on NYC dataset. RO-1 to RO-5 represent the variants of removing the following modules from the integrated RiskOracle in turn, PKDE strategy, ST-DFM, Overall affinity, Differential feature generator and Multi-task with HARS. The integrated model consistently outperforms other variants on both 30-minute and 10-minute levels. Specifically, the time-varying overall affinity and PKDE strategy contribute to the most remarkable promotions. We can conclude that the well-designed components exactly result in significant improvements in short-term predictions according to Table~\ref{tab:Ablation-NYC}.

\subsection{Hyper-parameter Studies}

Here we show the parameter studies on 30-minute level in NYC. We adjust the number of layers and filters in each layer to reach the best performance at 9 layers with 384 filters. We fix the weight of the main task as 1, and adjust $\lambda _1$, $\lambda _2$, Acc@$K$ arrives the highest 53.82\% when $\lambda_1$ = 0.8 and $\lambda_2$ = 1. Also, we adjust the weight of the dynamic element in overall affinity and reach the best performance when $\gamma$ equals $0.5$ among \{0, 0.5, 1.0, 1.2, 1.5\}. And $q$ equals 18 among \{9, 18, 33\} when the MSE of the second auxiliary task reaches the lowest. Note that our framework is trained offline and the parameters learned are utilized for online prediction. The computation workload can be determined by the parameters and done in several seconds, which sufficiently meets the real-time forecasting requirement.

\begin{table}[]\footnotesize
	\centering 
	\caption{Ablation studies on NYC dataset}
	\label{tab:Ablation-NYC} 
	\begin{tabular}{c|ccc|ccc}
		\toprule
		\hline
		{}& \multicolumn{3}{c|} {30-minute Interval}     &\multicolumn{3}{c} {10-minute Interval}     \\ \hline
		\tabincell{c}{Ablation\\Variants}  & MSE &\tabincell{c}{Acc\\@20}& \tabincell{c}{Acc\\@$K$}  & MSE & \tabincell{c}{Acc\\@6} &\tabincell{c}{Acc\\@$K$}   \\		
		RO-1 & 0.069  & 48\% & 42\% & 0.048 & 28\% & 25\%       \\ 
		RO-2 & 0.126  & 69\% & 53\% & 0.103 & 43\% & 38\%       \\
		RO-3 & 0.169  & 34\% & 36\%   & 0.124  & 29\% & 30\%      \\
		RO-4 & 0.115 & 63\% & 49\%  & 0.063 & 42\% & 31\%     \\
		RO-5 & 0.118 & 65\% & -               & 0.053 & 40\% & - \\
		\textbf{\tabincell{c}{Inte-\\grated}} & \textbf{0.108} & \textbf{70\%} & \textbf{57\%} & \textbf{0.045} & \textbf{45\%} & \textbf{46\%}   \\ \hline
		\bottomrule
		
	\end{tabular}
\end{table}

\subsection{Case Study}
\begin{figure}[!ht]	
	\centering
	\includegraphics[width=.95\columnwidth]{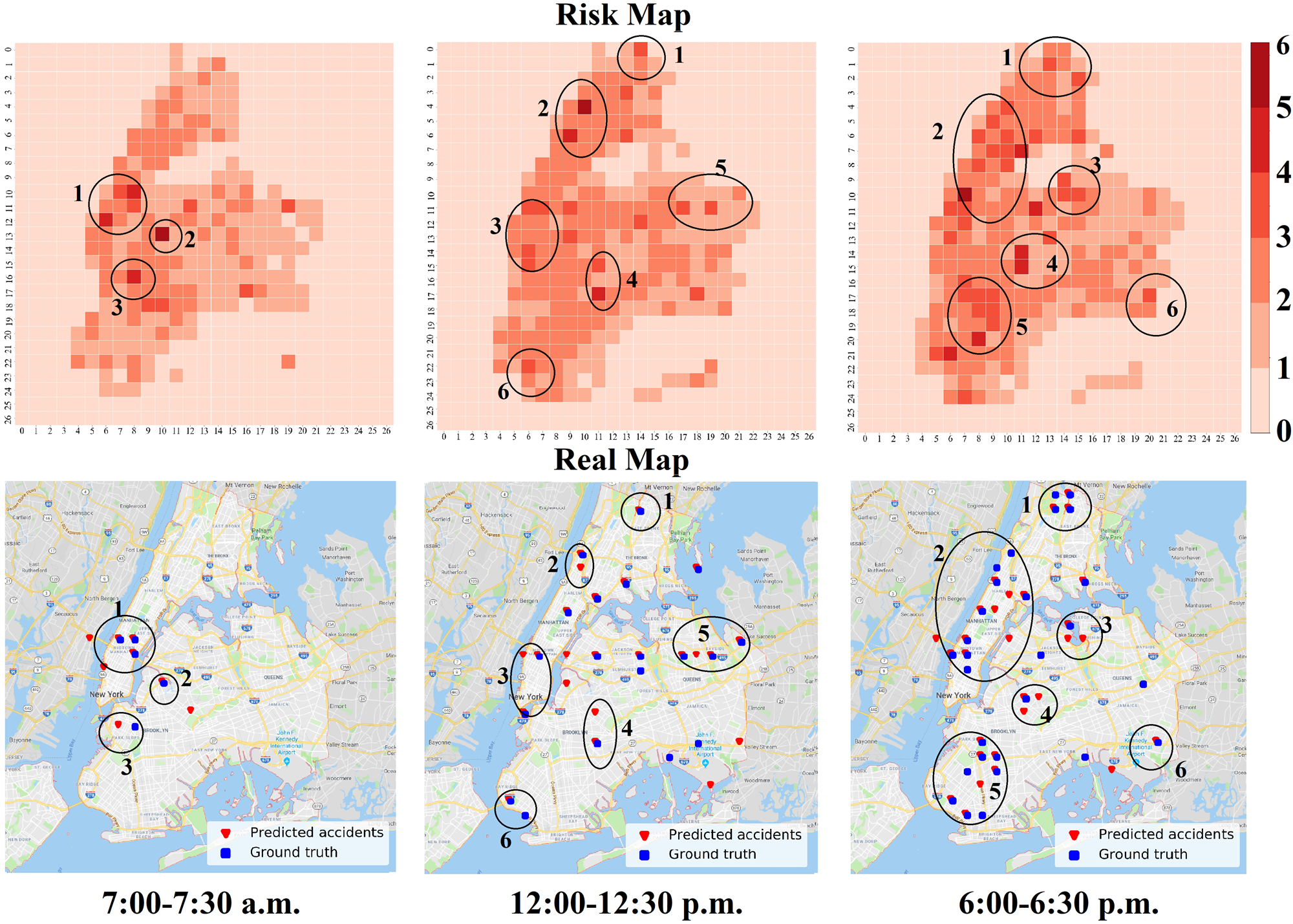}
	\caption{RiskOracle on May, 22th, 2017, NYC}
	\label{fig:Results_visual}

\end{figure}

We visualize the accidents predicted by RiskOracle at selected 30-minute intervals on one day in Figure~\ref{fig:Results_visual}. Overall, citywide risk maps generated by RiskOracle reveal discriminating risks and the highlighted subregions show great spatial similarities with the ground truth. Here, accidents predicted at 7:00 a.m. are rare, due to that few people go out on Sunday morning. However, the number of accidents increases when afternoon comes and it becomes even worse in the evening. It is mainly because of the heavy rain that evening, causing accident-prone road conditions. The results prove that the auxiliary task and HARS learn to adjust inferences accordingly by capturing dynamic patterns of accident distributions with external factors, which brings in better adaptivity than a unified threshold solution. 

\section{Conclusion}
In this paper, we tackle the challenges of minute-level citywide traffic accident forecasting by proposing the integrated framework RiskOracle based on Multi-task DTGN, providing a quantitive decision-making basis for urban safety in a more timely manner. We first propose two strategies to overcome the zero-inflated issue and sparse sensing. By incorporating the differential feature generator and time-varying overall affinity in Multi-task DTGN, our framework has the power to model sporadic spatiotemporal data and capture the short-term subregion-wise correlations. We also highlight most-likely accident regions to deal with spatial heterogeneities with learnable multi-scale accident distributions in the multi-task scheme. Experiments on two real-world datasets verify our framework outperforms the state-of-the-art solutions. Therefore, our work can be a paradigm for addressing spatiotemporal data mining tasks with sporadic labels and insufficient sensing data, e.g. predictions of the crimes and epidemic outbreaks.

\section{Acknowledgements}
This paper is partially supported by the Anhui Science Foundation for Distinguished Young Scholars (No.1908085J24), NSFC (No.61672487, No.61772492), Jiangsu Natural Science
Foundation (No.BK20171240, BK20191193) and CAS Pioneer Hundred Talents Program.

\small
\bibliography{1664_reference}
\bibliographystyle{aaai}

\end{document}